\documentclass[runningheads]{llncs}
\usepackage[T1]{fontenc}
\usepackage{graphicx}
\usepackage{booktabs}
\usepackage[misc]{ifsym}
\newcommand{\corr}{(\Letter)}
\usepackage{mwe}
\usepackage[hidelinks]{hyperref}
\hypersetup{
  bookmarks=true,
  bookmarksopen=true,
  bookmarksnumbered=true,
	bookmarksdepth=3,
}
\usepackage{threeparttable}
\usepackage{csquotes}
\usepackage{tabularx}
\usepackage{multirow}
\usepackage{makecell}
\usepackage[font=small]{caption} 
\usepackage{subcaption}
\usepackage[table]{xcolor}
\usepackage{fontawesome} 
\usepackage{pifont}
\usepackage[most]{tcolorbox}
\usepackage{etoolbox}
\robustify\tcbox
\usepackage{rotating}
\usepackage{longtable}
\usepackage{xltabular}
\usepackage{threeparttablex}
\AtBeginEnvironment{tablenotes}{\small} 
\usepackage{placeins}
\usepackage{soul}
\usepackage{xcolor}
\usepackage{enumitem}

\newbool{revision}
\boolfalse{revision}
\newcommand\revhl[1]{\ifbool{revision}{\hl{#1}}{#1}}
\newcommand{\rot}[1]{\rotatebox{90}{\makecell{#1}}}
\newcommand{\Yes}{\ding{51}}
\newcommand{\No}{\textcolor{black!20}{\ding{55}}}
\definecolor{gray}{RGB}{245,245,245}

\begin{document}
\title{Software Frameworks for Explainable AI in Time Series Classification: A Systematic Review}
\titlerunning{Software Frameworks for XAI in TSC: A Systematic Review}


\author{Louis Peter\inst{1,2}\orcidID{0009-0001-7478-1888} \and
Nils Gumpfer\inst{1,2}\orcidID{0000-0001-8644-9885} \and
Jana Fischer\inst{1,2}\orcidID{0009-0005-0445-0847} \and
Christin Seifert\inst{2,3}\orcidID{0000-0002-6776-3868} \and
Jennifer Hannig\inst{1,2}\orcidID{0000-0002-2789-5540} \corr}

\authorrunning{L. Peter et al.}

\institute{Technische Hochschule Mittelhessen - University of Applied Science, Friedberg, Germany
\and
Hessian Center for AI (hessian.AI), Darmstadt, Germany
\email{\{louis.peter,nils.gumpfer,jana.fischer,jennifer.hannig\}@kite.thm.de}
\and
Marburg University, Marburg, Germany
\email{christin.seifert@uni-marburg.de}}

\maketitle              

\begin{abstract}
Time series arise in a wide range of application domains and are analyzed using machine learning in decision-critical settings. Time series classification (TSC) is one of the most widely studied and relevant tasks. In this context, ensuring the transparency and trustworthiness of TSC models has become an important requirement, motivating the use of explainable artificial intelligence (XAI) methods.
Despite growing interest, research on XAI for TSC remains fragmented, and a systematic understanding of the available software frameworks for explanation generation, their evaluation practices, and practical limitations is still lacking. Prior work largely focused on individual explanation methods, while cross-framework consistency, time-series-specific evaluation, and reproducibility have received little attention.
In this survey, we analyze existing software frameworks for explanation generation and evaluation in TSC. We compare them along multiple dimensions, including supported XAI methods, evaluation metrics, usability, benchmarking support, and reproducibility, providing the first time-series-specific survey of frameworks with implementation comparisons and an analysis of frequency-domain support. We identify six frameworks that explicitly support time series and reveal common limitations: only one method supports frequency-domain explanations despite their relevance; only two evaluation metrics have been developed specifically for time series; and identical XAI methods can yield substantially different explanations across frameworks.
Based on these findings, we discuss open challenges and outline directions for future research, highlighting the need for unified, time-series-specific XAI frameworks that enable faithful, reproducible, and time-series-aware explanations.

\end{abstract}

\section{Introduction}
Time series data are a central part of many real-world applications in healthcare, industrial monitoring, finance, and audio processing. The increasing use of deep learning models for time series classification (TSC) has led to substantial gains in predictive performance, but these improvements often come at the cost of interpretability and transparency. In high-risk domains, understanding and justifying automated decisions is critical, both to foster user trust and to comply with regulatory requirements such as the European Union’s Artificial Intelligence Act (AI Act), in particular Article 14, which requires human oversight to ensure that users can interpret and effectively oversee AI outputs~\cite{eu_ai_act_2024}. These regulatory developments further highlight the necessity of reliable and faithful explanation methods for TSC.

Explainable artificial intelligence (XAI) has consequently emerged as an important research area. However, most existing XAI research has been developed and evaluated primarily in the context of image data~\cite{Nauta2023}. As a result, many explanation methods are applied to time series without being specifically adapted or systematically evaluated~\cite{Schlegel2019}, leaving their faithfulness and practical usefulness for time series data unclear.

Explanations for time series are less intuitive than for other data modalities, increasing the risk of misleading interpretations~\cite{Rojat2021}. While image classification tasks usually allow even non-experts to visually distinguish between classes such as cats and dogs, many real-world time series lack a clear notion of interpretable components. The raw time series signal often appears noisy and does not exhibit visually interpretable structure, even for domain experts. A representative example is an audio signal of a spoken digit, which is not visually interpretable in the time domain (see the left part of Fig.~\ref{fig:audio_mnist_frequency}). In such cases, explanations based solely on the time domain are often insufficient, and alternative representations in the frequency and time-frequency domains are usually considered. This highlights the importance  of incorporating these representations to generate meaningful explanations. 
\begin{figure}
    \centering
    \includegraphics[width=.76\linewidth]{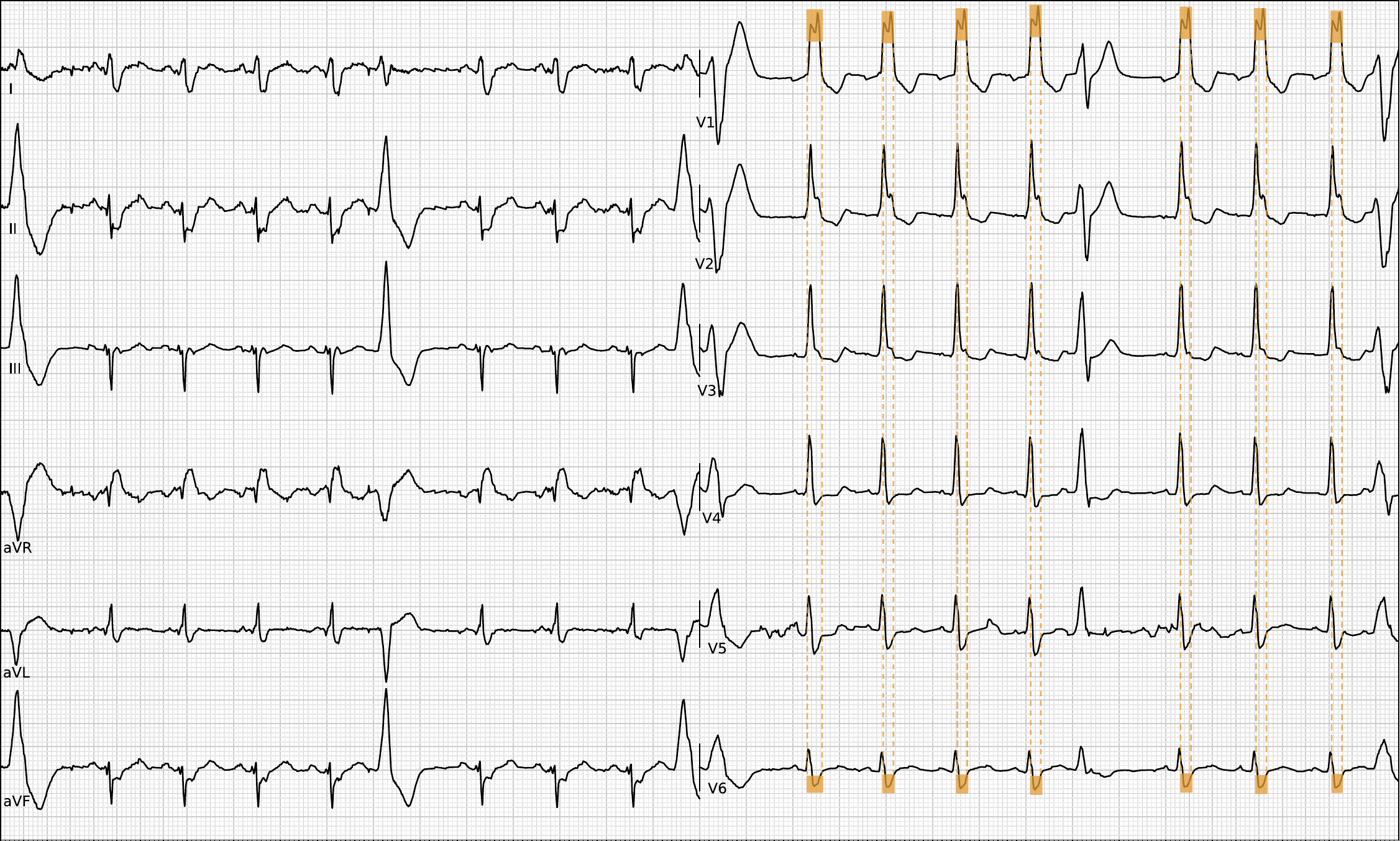}
    \caption{The electrocardiogram (ECG) signal is a multivariate time series in which each channel corresponds to a distinct measurement location (lead). All channels reflect the same underlying cardiac process with strong temporal and synchronous cross-channel dependencies. An example with right bundle branch block (RBBB) shows characteristic patterns in leads V1 and V6, highlighting the need for multichannel analysis.}
    \label{fig:ecg}
\end{figure}
In contrast, the electrocardiogram (ECG) is a multivariate time series for which meaningful and clinically interpretable patterns are well defined and visible in the time domain. Pathology-related patterns can be reliably interpreted by domain experts, particularly cardiologists, and often arise from complex cross-channel dependencies, reflecting spatial and temporal interactions of cardiac activity (see Fig.~\ref{fig:ecg}). Consequently, explaining multivariate time series requires XAI methods that explicitly account for cross-channel dependencies, which can be synchronous (i.e., dependencies between channels occurring at the same time step) or asynchronous.

Overall, time series pose unique challenges for explainability. Meaningful patterns often emerge from complex temporal dynamics and cross-channel dependencies, including interactions across time, frequency, or time-frequency domains. As a result, XAI for TSC differs fundamentally from image-based explanations and motivates the need for the systematic comparison, evaluation, and development of XAI methods tailored to the unique properties of time series.

Prior surveys have identified key challenges of XAI for time series data. Theissler et al.~\cite{Theissler2022} and Rojat et al.~\cite{Rojat2021} reviewed existing XAI methods and emphasized the need for systematic evaluation but focused primarily on methods rather than the frameworks\footnote{Throughout, we use ``framework'' to refer to a \emph{software} framework, i.e., a library or tool for generating and/or evaluating explanations.} used to implement, evaluate, and benchmark them in practice. Although multiple frameworks for evaluation exist, the survey by Le et al.~\cite{Le2023} showed that, as of 2022, only a single framework explicitly supported time series data. At the same time, Theissler et al.~\cite{Theissler2022} called for unified frameworks to enable comparative and reproducible evaluation of XAI methods for time series. It remains unclear how far this call has been addressed. In particular, there is no consolidated understanding of how recent XAI frameworks support TSC, how they differ in explanation generation and evaluation, or how they align with emerging regulatory requirements.

Addressing this gap motivates the present survey, which provides the first systematic review and comparative analysis of recent XAI frameworks for TSC. In contrast to prior surveys that focus on XAI methods~\cite{Theissler2022,Rojat2021} or analyze frameworks across different data types~\mbox{\cite{Le2023}}, we provide the first framework-centric analysis specific to XAI for TSC. We analyze how existing frameworks support the generation, evaluation, and benchmarking of explanations for TSC and assess their usability and practical limitations. 
\begin{enumerate}[label=RQ\arabic*:, leftmargin=*]
\item Which XAI frameworks currently support TSC, and how do they differ in usability and practical applicability? (Sec.~\ref{subsec:frameworks})
\item How comprehensively do these frameworks support explanation methods and evaluation metrics, and to what extent do they support univariate and multivariate time series? (Sec.~\ref{subsec:frameworks})
\item To what extent do current frameworks rely on reusing generic XAI methods and evaluation metrics, and how well do they support alternative signal representations? (Sec.~\ref{subsec:methods}-\ref{subsec:metrics})
\item How do existing frameworks support benchmarking of XAI methods for TSC, particularly with respect to dataset support and available ground-truth information? (Sec.~\ref{subsec:benchmark})
\item To what extent do differences in framework design and implementation affect the explanations and evaluation results produced by ostensibly identical XAI methods? (Sec.~\ref{subsec:reproducibility})
\end{enumerate}

\noindent
By addressing these questions, this survey provides a consolidated view of the current landscape of XAI frameworks for TSC, highlights systematic strengths and shortcomings, and identifies open challenges for robust, comparable, and time-series-aware XAI frameworks.

\section{Methodology}
\begin{figure}
    \centering
    \includegraphics[width=.85\linewidth]{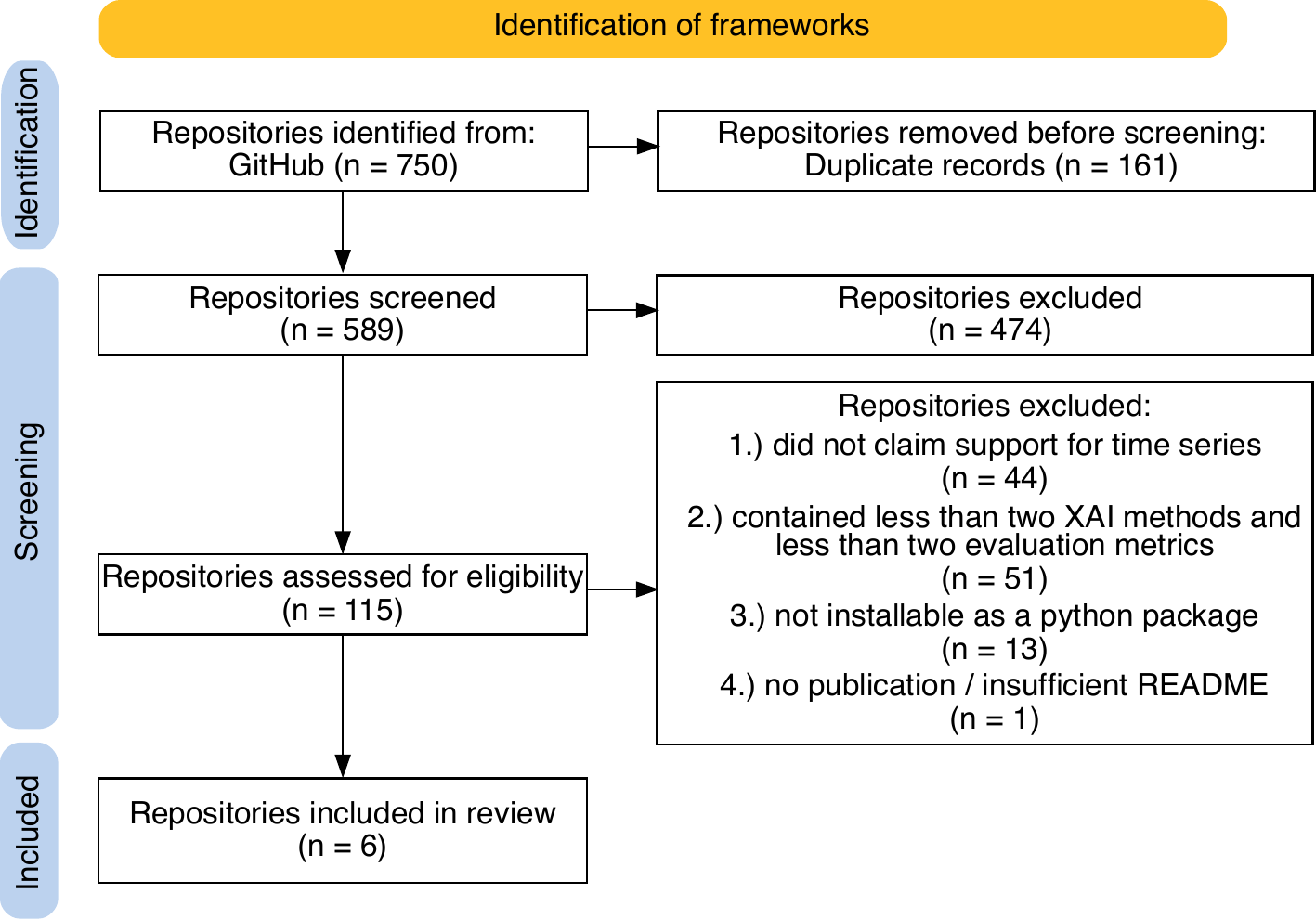}
    \caption{Flowchart of selection and review process.}
    \label{fig:prisma}
\end{figure}

We performed a systematic review of available XAI frameworks for TSC. In this work, an XAI framework denotes any software -- such as library or tool -- that facilitates either the comparative generation of explanations and/or the evaluation of XAI methods. The review process is illustrated in Fig.~\ref{fig:prisma}. We searched GitHub, which is the dominant platform for hosting open-source projects, using the following queries, resulting in 750 entries\footnote{The search was conducted on Dec 5th, 2025.}: \texttt{"explainable-ai time-series"}, \texttt{"explainable-ai evaluation"}, \texttt{"xai time-series"}, \texttt{"xai evaluation"}, and \texttt{"explanation time-series"}.

After removing duplicates, 589 repositories remained. To exclude unmaintained or incomplete repositories and ensure community relevance, we filtered repositories with fewer than four stars\footnote{GitHub stars represent the number of users who have marked a repository as a favorite.} and fewer than four forks\footnote{GitHub forks represent user copies of a repository to work on.} resulting in 115 repositories. We then manually assessed the remaining repositories and excluded those that 1.) did not claim the support of time series in the corresponding publication, README file, or documentation, \textit{or} 2.) contained less than two XAI methods and less than two evaluation metrics, \textit{or} 3.) were not installable as a python package, \textit{or} 4.) had no associated publication or an insufficient README file\footnote{A README was regarded insufficient if it provided neither a tutorial nor documentation for usage.}.

We identified six candidate repositories. We further examined the frameworks dependencies to identify additional frameworks, resembling a backward search for frameworks; however, all identified dependencies met at least one exclusion criterion. As a result, the final survey included six XAI frameworks.
For each framework, we conducted a systematic review of the source code repositories, associated publications, documentation, supported datasets, and -- where available -- the original publications describing the implemented XAI methods and evaluation metrics. The analysis was structured around the following dimensions:

\subsubsection{Frameworks.}
For each framework, we record the number of XAI methods and evaluation metrics, the supported machine-learning backends, whether it is available at a package index, and the corresponding GitHub repository metadata (last update, stars, and forks). To assess general usability, we adopted the usability scores proposed by Le et al.~\cite{Le2023}, which evaluate the usability of frameworks along three dimensions: \textit{active maintenance, interaction with community}, and \textit{documentation}. Each dimension is scored on a scale from 0 to 5 based on predefined criteria. 

\subsubsection{XAI Methods.}
We categorized the XAI methods into counterfactual, gradient-based, and perturbation-based approaches; methods that did not fit these categories were grouped under \textit{other}. We examined whether methods explicitly claim to support time series. For time-series-specific methods, we analyzed their applicability to univariate or multivariate time series and the domains in which explanations are generated (time, frequency, or time-frequency).
For methods implemented in multiple frameworks, we compared the resulting explanations to assess cross-framework reproducibility.

\subsubsection{Evaluation Metrics.}
We investigated whether evaluation metrics are compatible with time series (as explicitly stated) or specific to time series (i.e., originally developed for time series in the corresponding publication). For metrics specific to time series, we analyzed the domains in which explanations can be evaluated (time, frequency, or time-frequency). We categorized metrics into perturbation-based and ground-truth-based metrics; metrics that did not fit these categories were grouped under \textit{other}. For metrics implemented in multiple frameworks, we compared the resulting scores to assess reproducibility across implementations.

\subsubsection{Benchmarking Capabilities.}
We assessed whether frameworks support benchmark datasets for comparing and evaluating XAI methods. Datasets were considered if they can be accessed from the installed packages. We further examined whether the datasets provide ground-truth information (known or expected reference explanations) and whether they contain uni- or multivariate time series.

\begin{figure}[t]
\centering
\includegraphics[width=0.7\linewidth]{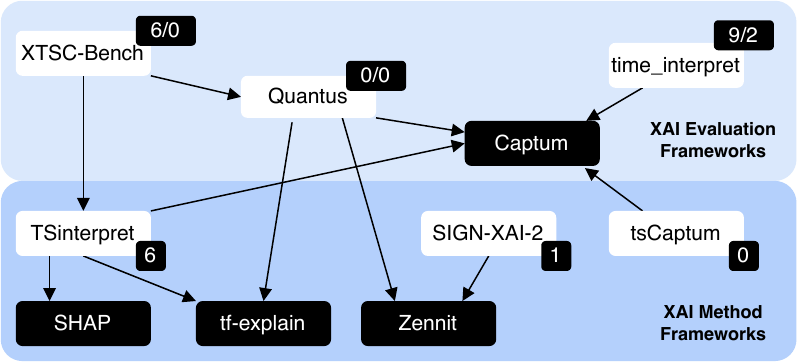}
    \caption{
    Dependency ecosystem of XAI frameworks (arrows indicate dependencies). White nodes support time series; black nodes are core dependencies without time series support. For XAI method frameworks, \tcbox[on line, colback=black!80, colframe=black, boxrule=0pt, arc=2pt,left=0pt, right=0pt, top=0pt, bottom=0pt]{\textcolor{white}{\tiny X}} indicates the number of time-series-specific XAI methods. For XAI evaluation frameworks, \tcbox[on line, colback=black!80, colframe=black, boxrule=0pt, arc=2pt,left=0pt, right=0pt, top=0pt, bottom=0pt]{\textcolor{white}{\scriptsize X/Y}} indicates the number of time-series-specific XAI methods (X) and evaluation metrics (Y).
    }
    \label{fig:dependencies}
\end{figure}

\section{Results}

We present our findings along the five questions introduced above. We first characterize the six frameworks and their usability (Sec.~\ref{subsec:frameworks}), then analyze XAI methods (Sec.~\ref{subsec:methods}), evaluation metrics (Sec.~\ref{subsec:metrics}), and benchmarking capabilities (Sec.~\ref{subsec:benchmark}), and finally assess cross-framework reproducibility (Sec.~\ref{subsec:reproducibility}).

\subsection{Frameworks}
\label{subsec:frameworks}
Of the six identified frameworks, three support both explanation generation and evaluation by providing XAI evaluation metrics in addition to XAI methods (time\_interpret~\cite{Enguehard2023}, Quantus~\cite{Hedstroem2023}, and XTSC-Bench~\cite{Hoellig2023}), while the remaining three focus exclusively on XAI methods (TSinterpret~\cite{Hoellig2022}, tsCaptum~\cite{Serramazza2024}, and SIGN-XAI-2~\cite{Gumpfer2023}). 
Table~\ref{tab:xai_frameworks} summarizes the frameworks and their main characteristics. All analyzed frameworks support PyTorch. TSInterpret has the broadest backend support, with native compatibility with PyTorch, scikit-learn, and TensorFlow; XTSC-Bench inherits these backends as it builds on TSInterpret. Usability scores (active maintenance, interaction with the community, and documentation) are consistently high, with mean values of (4.3, 3.6, 4.6) for XAI method frameworks and (4, 3.3, 5) for XAI evaluation frameworks. 
Quantus, which has the highest numbers of GitHub stars and forks, supports the largest set of XAI methods (26) by reusing or wrapping implementations from Captum~\cite{Kokhlikyan2020}, Zennit~\cite{Anders2021}, and tf-explain~\cite{Meudec2021}. However, none of these XAI methods are specific for time series (see Fig.~\ref{fig:dependencies}). Quantus also provides the largest collection of evaluation metrics (36). The largest set of time-series-specific XAI methods (9, all of which support multivariate time series) and metrics (2) is provided by time\_interpret. SIGN-XAI-2 extends Zennit, while tsCaptum wraps Captum and applies chunking to reduce computational complexity, despite not providing time-series-specific methods.

\begin{table*} [t!]
\begin{threeparttable}
\setlength{\tabcolsep}{2pt}
\centering
\caption{Overview of the frameworks with a reference to the papers proposing frameworks or to the software release (URLs are given in the reference entries).\faHistory: last update on GitHub;
\faStar: number of stars on GitHub;
\faCodeFork: number of forks on GitHub;
\faDownload: availability local (L) or as installable package (P);
\faCode: supported machine-learning backend (TensorFlow (T), PyTorch (P), and Scikit-learn (S)); 
\faBarChart: Usability scores in order of active maintenance, interaction with community, and documentation (3 scores denoted as X--X--X, each in the range 0--5 from lowest to best);
\faHashtag: number of supported XAI methods (implemented methods and, in parenthesis, wrapped or reused implementations of other frameworks);
\faLineChart: number of methods specific to univariate time series;
\faRandom: number of methods specific to multivariate time series;
\reflectbox{\faSignal}: number of methods generating explanations in the frequency domain;
\faTachometer: number of supported XAI metrics (implemented metrics and, in parenthesis, wrapped or reused implementations of other frameworks);
\faClockO: number of metrics specific to time series.}
\label{tab:xai_frameworks}
\begin{tabular*}{\textwidth}{lc|r|r|c|c|c|c|c|c|c|c|c}
\toprule
& \faHistory & \faStar & \faCodeFork  & \faDownload & \faCode & \faBarChart & \multicolumn{4}{|c|}{\textbf{XAI Methods}} & \multicolumn{2}{c}{\textbf{Metrics}}  \\
\midrule
\multicolumn{7}{l}{\textbf{XAI Method Frameworks}} &
\multicolumn{1}{|c|}{\faHashtag} & \faLineChart & \faRandom & \multicolumn{1}{c|}{\reflectbox{\faSignal}} & \faTachometer & \multicolumn{1}{c}{\faClockO}\\
\midrule
\rowcolor{gray}
TSInterpret~\cite{Hoellig2022} & 11/2025 & 144 & 18 & P & T, P, S & 5-4-5 & 6 + (10) & 2 & 4 & 0 & 0 & 0 \\
tsCaptum~\cite{Serramazza2024} & 11/2024 & 10 & 1 & P & P & 3-2-4 & (5) & 0 & 0 & 0 & 0  & 0 \\ 
\rowcolor{gray}
SIGN-XAI-2~\cite{Gumpfer2023} & 12/2025 & 6 & 0 & P & P & 5-5-5 & 2 + (9) & 1 & 0 & 1 & 0  & 0 \\
\midrule
\multicolumn{9}{l}{\textbf{XAI Evaluation Frameworks}}\\
\midrule
\rowcolor{gray}
Quantus~\cite{Hedstroem2023} & 07/2025 & 631 & 83 & P & T, P & 5-4-5 & (26) & 0 & 0 & 0 & 36  & 0 \\
time\_interpret~\cite{Enguehard2023} & 09/2025 & 71 & 10 & P & P & 5-4-5 & 18 & 0 & 9 & 0 & 17  & 2 \\
\rowcolor{gray}
XTSC-Bench~\cite{Hoellig2023} & 10/2023 & 4 & 0 & L & T, P, S & 2-2-5 & (16) & (2) & (4) & 0 & (11) & 0  \\ 
\bottomrule
\end{tabular*}
\end{threeparttable}
\end{table*}

\subsection{XAI Methods}
\label{subsec:methods}
The frameworks contain 51 XAI methods, including 16 perturbation-based, 25 gradient-based, 4 counterfactual, and 6 other methods (see Table~\ref{tab:methods} in the Appendix). Only 16 of these methods explicitly state that they were developed for time series, although the remaining methods may still be applicable. The majority of these 16 methods target multivariate time series, while only three are restricted to univariate time series (DFT-LRP~\cite{Vielhaben2024}, LEFTIST~\cite{Guilleme2019}, and NUN-CF~\cite{Delaney2021}). The multivariate methods are in principle also applicable to univariate signals (with two exceptions), whereas the three univariate-specific methods are not designed for the multivariate case. Two XAI methods can be used to adapt existing XAI methods to time series. Temporal Saliency Rescaling calculates two relevance scores~\cite{Ismail2020}: one for all time points and one for all features by masking them individually; the final explanation is the product of the two. Time Forward Tunnel (TFT) calculates the relevance for each time point using only the time points preceding it, disallowing the usage of future time points~\cite{Enguehard2023}. This can be used to force XAI methods to respect the temporal aspect of time series data. These methods can be options to apply XAI methods of other domains in TSC tasks while considering temporal structure.

\begin{figure}[t]
    \centering
    \includegraphics[width=\linewidth]{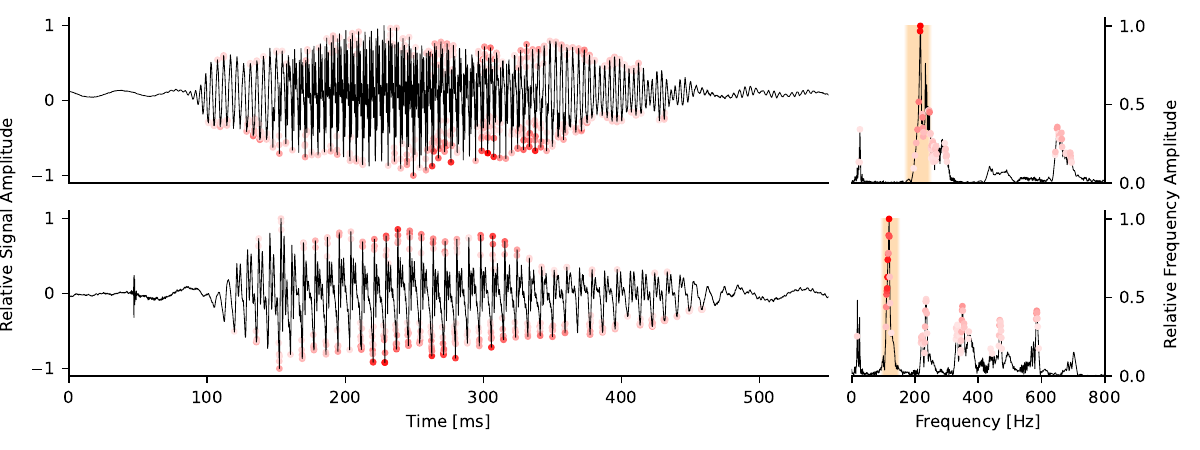}
    \caption{Spoken digit nine from the AudioMNIST data by a female (upper) and a male (lower) speaker. DFT-LRP explanations (SIGN-XAI-2 framework) are shown in the time (left) and frequency domain (right).
    Red dots indicate features (time steps or frequencies) with positive relevance scores; higher intensity corresponds to higher relevance. The expected ground truth in the frequency domain is highlighted in orange.}
    \label{fig:audio_mnist_frequency}
\end{figure}

For a time series as input, only a single method can generate explanations in the frequency and time-frequency domains. DFT-LRP combines Fourier transforms with Layer-wise Relevance Propagation (LRP)~\cite{Bach2015} to generate explanations for the time, frequency, and time-frequency domains and is developed for univariate data~\cite{Vielhaben2024}. The applied Fourier transforms can also be combined with other methods than LRP. In many TSC tasks, class-relevant patterns are present in multiple domains (time, frequency, and time-frequency). Fig.~\ref{fig:audio_mnist_frequency} illustrates an example of such a TSC task: the digit nine spoken by a female (Fig.~\ref{fig:audio_mnist_frequency} upper part) and by a male speaker (Fig.~\ref{fig:audio_mnist_frequency} lower part) from the AudioMNIST data set~\cite{Becker2024}. The classification task is to distinguish between female and male speakers.
We trained the model from~\cite{Vielhaben2024} using samples of the digit nine, with a balanced set of female and male speakers (same preprocessing as in~\cite{Vielhaben2024}, 1,200 records, 66.7/16.7/16.7\% train/validation/test split, 100\% test accuracy). We classified a single instance from a male and a female recording and generated explanations for the resulting predictions using DFT-LRP within the SIGN-XAI-2 framework. In the time domain (Fig.~\ref{fig:audio_mnist_frequency} left side), the explanations are scattered across the signal and concentrated on positive and negative peaks. In contrast, the frequency domain (Fig.~\ref{fig:audio_mnist_frequency} right side) clearly reveals the characteristic difference between a female and a male pronunciation of the digit nine. For adult speakers, the fundamental frequency typically ranges from 165 to 255~Hz for female voices, whereas male voices usually exhibit fundamental frequencies between 90 to 155~Hz~\cite{Baken2000}. The frequency regions corresponding to these ranges are highlighted in Fig.~\ref{fig:audio_mnist_frequency} and serve as ground-truth information. For both samples, the highest relevance scores are concentrated within their ground-truth regions.

\subsection{Evaluation Metrics}
\label{subsec:metrics}
The frameworks support a total of 52 evaluation metrics: 27 perturbation-based, 15 ground-truth-based, and 10 other (see Table~\ref{tab:metrics} in the Appendix). There is a small overlap in supported metrics, as only one metric is supported by all three XAI evaluation frameworks. XTSC-Bench builds entirely on Quantus for the implementation of its metrics, therefore, the metrics of XTSC-Bench are a subset of the metrics available in Quantus.
Out of the 52 evaluation metrics, 44 metrics are compatible with time series, six are not compatible (restricted to images) and only two are specifically developed for time series. The two metrics specific to time series are Mask Information and Mask Entropy~\cite{Crabbe2021}, they work on subsequences of time series measuring how well a predicted relevance fits to a ground-truth mask. None of the identified metrics are specifically developed for evaluating explanations within the frequency or time-frequency domain.

\subsection{Benchmarking Capabilities}
\label{subsec:benchmark}
Out of the six analyzed frameworks, three support datasets (XTSC-Bench, TSInterpret, and time\_interpret) for the comparative evaluation of explanations. The datasets available in each framework are listed in Table~\ref{tab:datasets}. XTSC-Bench offers six synthetic datasets proposed by Ismail et al.~\cite{Ismail2020}, which contain ground-truth information and are variable in length and number of channels. Additionally, it supports the University of California Riverside (UCR) and University of East Anglia (UEA) time series archives~\cite{Dau2019,Bagnall2018}, which are well established in AI research for TSC and comprise 128 and 30 datasets, respectively. Five datasets are supported by time\_interpret, including three synthetic datasets and two real-world clinical datasets (of which one is not openly accessible). The synthetic datasets are variable in length and number of channels and provide ground-truth information. The clinical datasets contain patient laboratory data and biomarkers but lack ground-truth information. The synthetic datasets of both XTSC-Bench and time\_interpret enable benchmarking of XAI methods for TSC due to their ground-truth information. TSInterpret also supports the UCR and UEA time series archives.
\begin{table} [t!]
    \centering
    \begin{threeparttable}
    \caption{Datasets of frameworks.
        \textbf{Frameworks (F)}: Frameworks supporting the dataset;
        \textbf{Channels and Length (L)}: Univariate (\faLineChart) or multivariate (\faRandom), synthetic datasets have variable (v) lengths and channel number;
        \textbf{Synthetic (S)}: \Yes~indicates dataset is synthetic, mixed (m), and \No~otherwise;
        \textbf{Ground Truth (GT)}: \Yes~indicates dataset contains GT information and \No~otherwise;
        *no open access; $^{\dagger}$Archives consist of multiple datasets; here we only indicate their variability at the collection level.}
    \label{tab:datasets}
    \begin{tabularx}{\linewidth}[t]{Xcccccc}
        \toprule
         \textbf{Dataset} & \textbf{Frameworks} & \multicolumn{2}{|c|}{\textbf{Channels}} & \textbf{L} & \textbf{S} & \textbf{GT} \\
         & & \multicolumn{1}{|c}{\faLineChart} & \multicolumn{1}{c|}{\faRandom} &  &  & \\
        \midrule
        \rowcolor{gray}
          Arma~\cite{Crabbe2021} & time\_interpret & \Yes & \Yes & v & \Yes & \Yes \\
          BioBank~\cite{Sudlow2015} & time\_interpret & * & * & * & * & * \\
        \rowcolor{gray}
          Hawkes~\cite{Bacry2018} & time\_interpret & \Yes & \Yes & v & \Yes & \Yes \\
          Hidden Markow Model~\cite{Crabbe2021} & time\_interpret & \Yes & \Yes  & v  & \Yes & \Yes \\
        \rowcolor{gray}
          Mimic III~\cite{Johnson2016} & time\_interpret & \No &  31 & 48 & \No & \No \\
          Gaussian, Harmonic, Pseudo Periodic, Autoregressive (AR), Continuous AR, Narma~\cite{Ismail2020} & time\_interpret & \Yes & \Yes  & v  & \Yes & \Yes  \\
        \rowcolor{gray}
         UCR Archive$^{\dagger}$~\cite{Dau2019} & \makecell[{{p{2cm}}}]{TSInterpret,\\XTSC-Bench} & \Yes & \No & v & m & \No \\
         UEA Archive$^{\dagger}$~\cite{Bagnall2018} & \makecell[{{p{2cm}}}]{TSInterpret,\\XTSC-Bench} & \No & \Yes  & v & m & \No \\
        \bottomrule
    \end{tabularx}
    \end{threeparttable}
\end{table}

\subsection{Reproducibility}
\label{subsec:reproducibility}
\begin{figure}[t]
    \centering
    \begin{subfigure}{0.20\columnwidth}
        \centering
        \includegraphics[width=.98\linewidth]{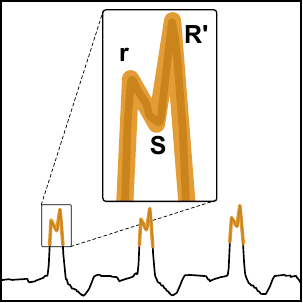}
    \end{subfigure}%
    \begin{subfigure}{0.40\columnwidth}
        \centering
        \includegraphics[width=.98\linewidth]{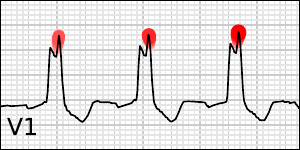}
    \end{subfigure}%
    \begin{subfigure}{0.40\columnwidth}
        \centering
        \includegraphics[width=.98\linewidth]{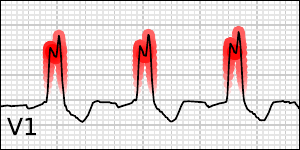}
    \end{subfigure}%
    \caption{Electrocardiogram (ECG) example with right bundle branch block (RBBB). Left: Illustration of characteristic \enquote{M}-shaped pattern (rSR'). Middle and right: Explanations generated using Integrated Gradients as implemented in TSInterpret (middle, based on Captum) and SIGN-XAI-2 (right, based on Zennit) for the same ECG instance, illustrating differences in relevance distributions across implementations. Relevance scores are indicated by red dots; higher intensity indicates higher relevance.}
    \label{fig:ig_comparison}
\end{figure}

We compared explanations and evaluation scores produced by XAI methods and evaluation metrics with different implementations across frameworks.
Five XAI methods (Saliency, Occlusion, Feature Ablation, Deconvolution, and Integrated Gradients, others were excluded due to different parameter usage or different backend support) were compared on a convolutional neural network (CNN) with the architecture described in Gumpfer et al.~\cite{Gumpfer2024}, trained for 20 epochs on right bundle branch block (RBBB) records and an equal amount of healthy ECG records of the PTB-XL dataset~\cite{Wagner2020} (full-length 12-lead ECGs, no preprocessing, 3,316 records, 80/10/10\% train/validation/test split, 96.98\% test accuracy). We explained the model predictions on the test set using Integrated Gradients (IG)~\cite{Sundararajan2017} as it is implemented in TSInterpret via Captum and in SIGN-XAI-2 via Zennit. IG attributes relevance by integrating gradients along a straight-line path between a baseline and the input. To ensure comparability, we used identical parameters in both frameworks (zero baseline and 50 interpolation steps). One example of resulting explanations is shown in Fig.~\ref{fig:ig_comparison}, illustrating one channel (lead V1) of the ECG example from Fig.~\ref{fig:ecg}. The relevance distributions differ noticeably and only the Zennit-based implementation (Fig.~\ref{fig:ig_comparison}, right) explains the expected \enquote{M}-shaped pattern characteristic of RBBB~\cite{Surawicz2009}. To verify the results, we analyzed the whole test set of 443 instances: Captum- and Zennit-based IG yielded a mean Spearman rank correlation of $-0.14$ (median $-0.17$, std 0.11; 0 of 443 instances had $\rho > 0.5$) and a top-5\% Jaccard overlap of 0.26 ($\pm0.09$). These results confirm that the divergence in Fig.~\ref{fig:ig_comparison} reflects systematic disagreement rather than an instance-specific or thresholding artifact. The other investigated methods provided identical explanations across frameworks, when configured with matching parameters. We also compared evaluation metrics across frameworks, but only Area under the Receiver Operating Characteristic Curve (ROC-AUC)~\cite{Fawcett2006} is implemented by two (Quantus and time\_interpret). Unlike the explanations, the metric produced consistent values across both implementations.

\FloatBarrier

\section{Open Challenges \& Further Research}
Compared to 2022, when only one framework supported time series data~\cite{Le2023}, we now identified five additional frameworks that claim to support time series, indicating increased research activity. This development suggests that earlier calls for time-series-aware XAI frameworks~\cite{Theissler2022} have been acknowledged by the community. However, our analysis shows that explicit support for time-series-specific properties remains limited.

\subsubsection{Method Applicability.}
Although many supported XAI methods are technically applicable to TSC, only a small fraction has been explicitly developed for time series. Two frameworks, Quantus and tsCaptum, do not include any XAI methods that are specifically designed for time series, but instead rely exclusively on reusing or wrapping methods developed for other data types. The remaining frameworks also partially rely on generic explanation methods. However, the suitability of the underlying methods for time series is often not explicitly validated. Further research should investigate to which extent these methods account for time-series-specific properties like temporal and cross-channel dependencies.

\subsubsection{Cross-Channel Dependencies.}
While most supported XAI methods can highlight relevant regions in multiple channels, the dependencies of these relevant parts are not represented and would require an additional dimension of explanation. This issue also manifests in the common use of heatmap-based relevance visualizations, which are unable to highlight class-relevant synchronous and asynchronous cross-channel dependencies. PAX-TS~\cite{Kreuzer2025}, a perturbation-based method, addresses this for forecasting tasks, but at higher computational cost. 
More generally, gradient-based approaches are efficient for high-dimensional inputs yet cannot explain cross-channel dependencies, whereas perturbation-based methods can, at increased cost, suggesting potential benefits of hybrid approaches for future research.

\subsubsection{Frequency and Time-Frequency Domains.} 
Another major limitation is the insufficient support for frequency and time-frequency domain explanations. In many TSC tasks, including audio, biomedical, and sensor-based applications, time-domain explanations alone are often insufficient, as class-relevant patterns are present in multiple domains (time, frequency, and time-frequency). DFT-LRP is currently the only XAI method of the analyzed frameworks that generates explanations in multiple domains for models using time-domain inputs. However, Fourier transformations, used by DFT-LRP, can also be combined with other XAI methods to extend their explanations to the frequency and time-frequency domains. Currently, among the analyzed frameworks only SIGN-XAI-2 implements a frequency-aware method, which remain rare in the literature as well.

\subsubsection{Evaluation Metrics.} 
Similar limitations exist for evaluation metrics. Although many metrics implemented in current frameworks are technically applicable to time series, it is often unclear to which extent they account for time-series-specific properties. An assumption that some metrics make is the independence of feature relevance, which is violated by temporal and cross-channel dependencies in time series. Only two of the 52 analyzed metrics were developed specifically for time series, in contrast to the literature, which offers time-series-specific metrics such as the Attribution Stability Indicator~\cite{Schlegel2023a}, Swap Time Points, and Mean Time Points~\cite{Schlegel2019}. Representations of time series in the frequency domain are not considered by any of the evaluation metrics. The application of existing perturbation-based metrics to evaluate explanations in the frequency domain is problematic, because small perturbations can impact the entire signal in the time domain. Future research should investigate to which extent supported metrics account for time-series-specific properties like temporal and cross-channel dependencies as well as frequency and time-frequency evaluations. We encourage researchers to develop dedicated metrics for these properties.

\subsubsection{Benchmarking and Ground-Truth Availability.}
Only half of the frameworks support benchmarking datasets and only two include ground-truth information. All datasets with ground truth are based on synthetic or predefined data distributions. It remains unclear to what extent they represent real time series. Extending existing real-world datasets with ground-truth annotations would allow for more reliable benchmarking and transfer to real applications.

\subsubsection{Reproducibility and Consistency.}
Our reproducibility analysis shows that explanation results may depend on the choice of the XAI framework. We observed substantial differences across frameworks for explanations generated by IG. For one framework, the class-relevant pattern was not visible in the resulting explanations. Prior work by Le et al.~\cite{Le2023} has shown that evaluation metrics can exhibit substantial variability across implementations. In contrast, we did not observe discrepancies for evaluation metrics in our experiments. However, this observation is likely limited due to the small overlap of evaluation metrics across frameworks, which restricts the scope of reproducibility analysis for metrics. 
We emphasize that researchers need to take care when comparing or reproducing published results (without knowledge of the used framework), this highlights current limitations of XAI research in generating reliable explanations.

\subsubsection{Transferability.}
Although this survey focuses on TSC, many of the identified limitations may also affect other time series tasks, such as forecasting and anomaly detection, which are even less studied within current XAI research for time series despite their practical relevance.

\subsubsection{Limitations.}
The systematic search underlying this review was conducted entirely on GitHub, which allowed us to specifically filter for software contributions. We acknowledge that this procedure might miss research software published on other platforms. Furthermore, our cross-framework reproducibility analysis is limited in scope: only five explanation methods and a single evaluation metric were implemented in more than one framework. Our results therefore demonstrate that explanations can depend on the framework implementation, but do not establish how prevalent this is across methods and metrics.

\section{Conclusion}
We presented the first systematic survey of XAI frameworks for TSC and identified substantial limitations in the current landscape. These include limited support for time-series-specific XAI methods and evaluation metrics, minimal incorporation of alternative signal representations such as frequency and time-frequency domains for XAI methods, a lack of ground-truth annotations for real-world benchmark datasets, and a lack of methods for explaining cross-channel dependencies in multivariate time series. We argue that researchers should exercise caution when applying methods or evaluation metrics for TSC from existing frameworks, as different implementations can yield substantially different results.
We provide Tables~\ref{tab:xai_frameworks} and~\ref{tab:datasets} as references to select frameworks and datasets for TSC tasks (Tables~\ref{tab:methods} and~\ref{tab:metrics} in the Appendix can serve as additional resources for selecting time-series specific methods and metrics).

Publicly funded, community-driven efforts in other disciplines, such as the platform Galaxy~\cite{galaxy2024} in bioinformatics, highlight the potential of shared infrastructure for reproducible and collaborative research. A standardized, community-driven initiative would greatly benefit XAI research by enabling faithful, reproducible, and time-series-aware explanations as well as robust, systematic comparison and evaluation of XAI methods.

\begin{credits}

\subsubsection*{\ackname}
This work was supported by the German Federal Ministry of Research, Technology and Space (BMFTR) through ExperTeam4KI (grant no.~16IS24063). We gratefully acknowledge support from the hessian.AI Service Center (funded by the BMFTR, grant no.~16IS22091) and the hessian.AI Innovation Lab (funded by the Hessian Ministry for Digital Strategy and Innovation, grant no.~S-DIW04/0013/003). This work was partially funded by the Deutsche Forschungsgemeinschaft (DFG, German Research Foundation) under project ID 536124560.

\subsubsection*{\discintname}
The authors have no competing interests to declare that are
relevant to the content of this article.
\end{credits}

%
%
%
\bibliographystyle{splncs04}
\bibliography{references}

\clearpage
\section*{Appendix A}
\addcontentsline{toc}{section}{Appendix A}
\renewcommand{\thetable}{A.\arabic{table}}
\setcounter{table}{0}
\setlength{\LTpre}{0pt}
\begin{ThreePartTable}
\begin{xltabular}{\linewidth}{>{\cellcolor{white}}c >{\raggedright\arraybackslash}X c c l c c c c c c}

\caption{XAI methods of frameworks.
    \textbf{Compatibility}: \Yes~indicates method is specific for time series,  otherwise marked as -- (not explicitly stated);
    \textbf{Dimensionality}:  supports uni- (U), multivariate (M) time series or both (B);
    \textbf{Frequency}: \Yes~generates explanations for the frequency and time-frequency domain;
    \textbf{Frameworks}: \Yes~indicates that the framework implements the method, (\Yes) if the framework wrapped or reused implementations of other frameworks, and \No~otherwise.
    Abbr.: Counterfactuals (CF)}

\label{tab:methods} \\

\toprule
& &
& & & \multicolumn{6}{c}{\textbf{Frameworks}} \\
\cmidrule(lr){6-11}
& \textbf{\Large Methods}
& \rot{\textbf{Compatibility}}
& \rot{\textbf{Dimensionality}}
& \rot{\textbf{Frequency}}
& \rot{\textbf{Quantus}}
& \rot{\textbf{TSInterpret}}
& \rot{\textbf{time\underline{\enspace}interpret}}
& \rot{\textbf{tsCaptum}}
& \rot{\textbf{XTSC-Bench}}
& \rot{\textbf{SIGN-XAI-2}} \\
\midrule
\endfirsthead

\multicolumn{11}{c}{\tablename~\thetable~-- \textit{continued from previous page}} \\[2pt]
\toprule
& &
& & & \multicolumn{6}{c}{\textbf{Frameworks}} \\
\cmidrule(lr){6-11}
& \textbf{\Large Methods}
& \rot{\textbf{Compatibility}}
& \rot{\textbf{Dimensionality}}
& \rot{\textbf{Frequency}}
& \rot{\textbf{Quantus}}
& \rot{\textbf{TSInterpret}}
& \rot{\textbf{time\underline{\enspace}interpret}}
& \rot{\textbf{tsCaptum}}
& \rot{\textbf{XTSC-Bench}}
& \rot{\textbf{SIGN-XAI-2}} \\
\midrule
\endhead

\midrule
\multicolumn{11}{r}{\textit{Continued on next page}} \\
\endfoot

\bottomrule
\endlastfoot

\rowcolor{gray}
    & BetaSmooth                                             & {-}  & {-} & \No  & \No    & \No    & \No  & \No & \No    & (\Yes) \\
    & Conductance                                            & {-}  & {-} & \No  & (\Yes) & \No    & \No  & \No & \No    & \No    \\
\rowcolor{gray}
    & Deconvolution                                          & {-}  & {-} & \No  & (\Yes) & \No    & \No  & \No & \No    & (\Yes) \\
    & Deep Learning Important Features (DeepLIFT)           & {-}  & {-} & \No  & (\Yes) & (\Yes) & \No  & \No & (\Yes) & \No    \\
\rowcolor{gray}
    & DeepLIFT Shapley Additive Explanations (SHAP)         & {-}  & {-} & \No  & (\Yes) & (\Yes) & \No  & \No & (\Yes) & \No    \\
    & DFT-LRP                                               & \Yes & U   & \Yes & \No    & \No    & \No  & \No & \No    & \Yes   \\
\rowcolor{gray}
    & Discretized Integrated Gradients                      & {-}  & {-} & \No  & \No    & \No    & \Yes & \No & \No    & \No    \\
    & Excitation Backpropagation                            & {-}  & {-} & \No  & \No    & \No    & \No  & \No & \No    & (\Yes) \\
\rowcolor{gray}
    & GeodesicIntegratedGradients                           & {-}  & {-} & \No  & \No    & \No    & \Yes & \No & \No    & \No    \\
    & Gradient-weighted Class Activation Mapping (Grad-CAM) & {-}  & {-} & \No  & (\Yes) & (\Yes) & \No  & \No & (\Yes) & \No    \\
\rowcolor{gray}
    & Gradient $\times$ Input                               & {-}  & {-} & \No  & (\Yes) & \No    & \No  & \No & \No    & \No    \\
    & GradientSHAP                                          & {-}  & {-} & \No  & (\Yes) & (\Yes) & \No  & \No & (\Yes) & \No    \\
\rowcolor{gray}
   & Guided Backpropagation                                & {-}  & {-} & \No  & (\Yes) & \No    & \No  & \No & \No    & (\Yes) \\
    & Guided Grad-CAM                                       & {-}  & {-} & \No  & (\Yes) & \No    & \No  & \No & \No    & \No    \\
\rowcolor{gray}
    & Integrated Gradients                                  & {-}  & {-} & \No  & (\Yes) & (\Yes) & \No  & \No & (\Yes) & (\Yes) \\
    & InternalInfluence                                     & {-}  & {-} & \No  & (\Yes) & \No    & \No  & \No & \No    & \No    \\
\rowcolor{gray}
    & Layer Activation                                      & {-}  & {-} & \No  & (\Yes) & \No    & \No  & \No & \No    & \No    \\
    & Layer-wise Relevance Propagation (LRP)                & {-}  & {-} & \No  & (\Yes) & \No    & \No  & \No & \No    & (\Yes) \\
\rowcolor{gray}
    & Saliency (Vanilla Gradients)                          & {-}  & {-} & \No  & (\Yes) & (\Yes) & \No  & \No & (\Yes) & (\Yes) \\
    & Sequential Integrated Gradients                       & {-}  & {-} & \No  & \No    & \No    & \Yes & \No & \No    & \No    \\
\rowcolor{gray}
    & SIGN                                                  & {-}  & {-} & \No  & \No    & \No    & \No  & \No & \No    & \Yes   \\
    \multirow{-25}{*}{\rotatebox[origin=c]{90}{\textbf{Gradient-based}}}  
    & SmoothGrad                                            & {-}  & {-} & \No  & (\Yes) & (\Yes) & \No  & \No & (\Yes) & (\Yes) \\
\rowcolor{gray}
    & Testing with Concept Activation Vectors (TCAV)        & {-}  & {-} & \No  & (\Yes) & \No    & \No  & \No & \No    & \No    \\
    & Temporal Integrated Gradients                         & \Yes & B   & \No  & \No    & \No    & \Yes & \No & \No    & \No    \\
\rowcolor{gray}
    & Tracing Gradient Descent with Checkpoints (TracInCP) & {-}  & {-} & \No  & (\Yes) & \No    & \No  & \No & \No    & \No    \\
\midrule
\rowcolor{gray}
    & Augmented Occlusion                                   & \Yes & B   & \No & \No    & \No    & \Yes & \No    & \No    & \No   \\
    & BayesKernelSHAP                                       & {-}  & {-} & \No & \No    & \No    & \Yes & \No    & \No    & \No   \\
\rowcolor{gray}
    & BayesLIME                                             & {-}  & {-} & \No & \No    & \No    & \Yes & \No    & \No    & \No   \\
    & DynaMask                                              & \Yes & B   & \No & \No    & \No    & \Yes & \No    & \No    & \No   \\
\rowcolor{gray}
    & Extremal Mask                                         & \Yes & B   & \No & \No    & \No    & \Yes & \No    & \No    & \No   \\
    & Feature Ablation                                      & {-}  & {-} & \No & (\Yes) & (\Yes) & \Yes & (\Yes) & (\Yes) & \No   \\
\rowcolor{gray}
    & Feature Permutation                                   & {-}  & {-} & \No & (\Yes) & \No    & \No  & (\Yes) & \No    & \No   \\
    & KernelSHAP                                            & {-}  & {-} & \No & (\Yes) & \No    & \No  & (\Yes) & \No    & \No   \\
\rowcolor{gray}
    & LEFTIST                                               & \Yes & U   & \No & \No    & \Yes   & \No  & \No    & (\Yes) & \No   \\
    & Local Interpretable Model-agnostic Explanations (LIME)& {-}  & {-} & \No & (\Yes) & \No    & \No  & (\Yes) & \No    & \No   \\
\rowcolor{gray}
    & Local Outlier Factor (LOF)-LIME                       & {-}  & {-} & \No & \No    & \No    & \Yes & \No    & \No    & \No   \\
    & LOF-KernelSHAP                                        & {-}  & {-} & \No & \No    & \No    & \Yes & \No    & \No    & \No   \\
\rowcolor{gray}
    & Occlusion                                             & {-}  & {-} & \No & (\Yes) & (\Yes) & \Yes & \No    & (\Yes) & (\Yes)\\
    & Shapley Value Sampling                                & {-}  & {-} & \No & (\Yes) & (\Yes) & \No  & (\Yes) & (\Yes) & \No   \\
\rowcolor{gray}
    & Temporal Augmented Occlusion                          & \Yes & B   & \No & \No    & \No    & \Yes & \No    & \No    & \No   \\
    \multirow{-17}{*}{\rotatebox[origin=c]{90}{\textbf{Perturbation-based}}}
    & Temporal Occlusion                                    & \Yes & B   & \No & \No    & \No    & \Yes & \No    & \No    & \No   \\
\midrule
\rowcolor{gray}
    & COMTE                                       & \Yes & M & \No & \No & \Yes & \No & \No & (\Yes) & \No \\
    & NUN-CF                                      & \Yes & U & \No & \No & \Yes & \No & \No & (\Yes) & \No \\
\rowcolor{gray}
    & Shapelet explainer for time series (SETS)   & \Yes & B & \No & \No & \Yes & \No & \No & (\Yes) & \No \\
    \multirow{-4}{*}{\rotatebox[origin=c]{90}{\textbf{CF}}}
    & TSEvo                                       & \Yes & B & \No & \No & \Yes & \No & \No & (\Yes) & \No \\

\midrule
\rowcolor{gray}
    & Activations Visualization                   & {-}  & {-} & \No & \No & \No  & \No  & \No & \No    & \No \\
    & Feature Importance in Time (FIT)            & \Yes & B   & \No & \No & \No  & \Yes & \No & \No    & \No \\
\rowcolor{gray}
    & Layer/Neuron Activations                    & {-}  & {-} & \No & \No & \No  & \No  & \No & \No    & \No \\
    & Reverse Time Attention model (RETAIN)       & \Yes & B   & \No & \No & \No  & \Yes & \No & \No    & \No \\
\rowcolor{gray}
    & Time Forward Tunnel                         & \Yes & B   & \No & \No & \No  & \Yes & \No & \No    & \No \\
    \multirow{-6}{*}{\rotatebox[origin=c]{90}{\textbf{Other}}}
    & Temporal Saliency Rescaling                 & \Yes & M   & \No & \No & \Yes & \No  & \No & (\Yes) & \No \\
\end{xltabular}
\end{ThreePartTable}
\begin{ThreePartTable}
	\setlength{\tabcolsep}{2pt}
	\begin{TableNotes}[para,flushleft]
			\item[1] Two metrics share the name \textit{Sufficiency} and are differentiated by the reference to their original publications.
	\end{TableNotes}
	\begin{xltabular}{\linewidth}[t]{>{\cellcolor{white}}c>{\raggedright\arraybackslash}Xcccc}
	\caption{Evaluation metrics of frameworks. Legend: \textbf{Compatibility}: \Yes~indicates metric is specific for time series, C if the metric is compatible with time series, and \No~otherwise; \textbf{Quantus, time\underline{\enspace}interpret, XTSC-Bench}: \Yes~indicates that the framework implements the metrics, (\Yes) if the framework wrapped or reused implementations of other frameworks, and \No~otherwise.}
	\label{tab:metrics}\\
        \toprule
        & \textbf{\Large Metrics} &
        \rot{\textbf{Compatibility}} &
        \rot{\textbf{Quantus}} &
        \rot{\textbf{time\underline{\enspace}interpret}} &
        \rot{\textbf{XTSC-Bench}} \\
				\midrule
				\endfirsthead

				\multicolumn{6}{c}{\tablename~\thetable~-- \textit{continued from previous page}} \\[2pt]
				\toprule
				& \textbf{\Large Metrics} &
        \rot{\textbf{Compatibility}} &
        \rot{\textbf{Quantus}} &
        \rot{\textbf{time\underline{\enspace}interpret}} &
        \rot{\textbf{XTSC-Bench}} \\
				\midrule
				\endhead
				\midrule
				\multicolumn{6}{r}{\textit{Continued on next page}} \\
				\endfoot
				\bottomrule
				\insertTableNotes
				\endlastfoot
        \rowcolor{gray}
        & Accuracy                         & C & \No  & \Yes & \No    \\
        \rowcolor{white}
        & Avg-Sensitivity                  & C & \Yes & \No  & (\Yes) \\
        \rowcolor{gray}
        & Comprehensiveness                & C & \No  & \Yes & \No    \\
        \rowcolor{white}
        & Continuity                       & \No & \Yes & \No  & \No    \\
        \rowcolor{gray}
        & Consistency                      & C & \Yes & \No  & \No    \\
        \rowcolor{white}
        & Cross\_entropy                   & C & \No  & \Yes & \No    \\
        \rowcolor{gray}
        & Faithfulness Correlation         & C & \Yes & \No  & (\Yes) \\
        \rowcolor{white}
        & Faithfulness Estimate            & C & \Yes & \No  & (\Yes) \\
        \rowcolor{gray}
        & Infidelity                       & \No & \Yes & \No  & \No    \\
        \rowcolor{white}
        & Iterative Removal of Features (IROF)& \No & \Yes & \No  & \No  \\
        \rowcolor{gray}
        & Lipschitz\_max                   & C & \No  & \Yes & \No    \\
        \rowcolor{white}
        & Local Lipschitz Estimate         & C & \Yes & \No  & \No    \\
        \rowcolor{gray}
        & Log\_odds                        & C & \No  & \Yes & \No    \\
        \rowcolor{white}
        & Max-Sensitivity                  & C & \Yes & \No  & (\Yes) \\
        \rowcolor{gray}
        & Mean absolute error (MAE)        & C & \No  & \Yes & \No    \\
        \rowcolor{white}
        & Mean squared error (MSE)         & C & \No  & \Yes & \No    \\
        \rowcolor{gray}
        & Monotonicity (Arya)              & C & \Yes & \No  & (\Yes) \\
        \rowcolor{white}
        & Monotonicity (Nguyen)            & C & \Yes & \No  & (\Yes) \\
        \rowcolor{gray}
        & Pixel Flipping                   & C & \Yes & \No  & \No    \\
        \rowcolor{white}
        & Region Perturbation              & C & \Yes & \No  & \No    \\
        \rowcolor{gray}
        & Relative Input Stability         & C & \Yes & \No  & \No    \\
        \rowcolor{white}
        & Relative Output Stability        & C & \Yes & \No  & \No    \\
        \rowcolor{gray}
        & Relative Representation Stability& C & \Yes & \No  & \No    \\
        \rowcolor{white}
        & Remove and Debias (ROAD)         & \No & \Yes & \No  & \No    \\
        \rowcolor{gray}
      	& Selectivity                      & \No & \Yes & \No  & \No    \\
        \rowcolor{white}
        & SensitivityN                     & C & \Yes & \No  & \No    \\
        \rowcolor{gray}
        \multirow{-27}{*}{\rotatebox[origin=c]{90}{\textbf{Perturbation-based}}}
        & Sufficiency\tnote{1}~~by DeYoung et al.~\cite{DeYoung2020}& C & \No  & \Yes & \No    \\
        \rowcolor{white}\pagebreak
        & Attribution Localisation         & C & \Yes & \No  & \No    \\
        \rowcolor{gray}
        & Area under precision curve (AUP) & C & \No  & \Yes & \No    \\
        \rowcolor{white}
        & Area under precision-recall curve (AUPRC)& C & \No  & \Yes & \No    \\
        \rowcolor{gray}
        & Area under recall curve (AUR)    & C & \No  & \Yes & \No    \\
        \rowcolor{white}
        & Area under the Receiver Operating Characteristic Curve (ROC-AUC)& C & \Yes & \Yes & (\Yes) \\
        \rowcolor{gray}
        & Focus                            & \No & \Yes & \No  & \No    \\
        \rowcolor{white}
        & MAE                              & C & \No  & \Yes & \No    \\
        \rowcolor{gray}
        & Mask entropy                     & \Yes & \No  & \Yes & \No    \\
        \rowcolor{white}
        & Mask information                 & \Yes & \No  & \Yes & \No    \\
        \rowcolor{gray}
        & MSE                              & C & \No  & \Yes & \No    \\
        \rowcolor{white}
        & Pointing Game                    & C & \Yes & \No  & (\Yes) \\
        \rowcolor{gray}
        & Relevance Mass Accuracy          & C & \Yes & \No  & (\Yes) \\
        \rowcolor{white}
        & Relevance Rank Accuracy          & C & \Yes & \No  & (\Yes) \\
        \rowcolor{gray}
        & Root mean squared error (RMSE)   & C & \No  & \Yes & \No    \\
        \rowcolor{white}
        \multirow{-16}{*}{\rotatebox[origin=c]{90}{\textbf{Ground-Truth-based }}}
        & Top-K Intersection               & C & \Yes & \No  & \No    \\
        \midrule
        \rowcolor{white}
        & Complexity                       & C & \Yes & \No  & (\Yes) \\
        \rowcolor{gray}
        & Effective Complexity             & C & \Yes & \No  & \No    \\
        \rowcolor{white}
        & Efficient MPRT                   & C & \Yes & \No  & \No    \\
        \rowcolor{gray}
        & Input Invariance                 & C & \Yes & \No  & \No    \\
        \rowcolor{white}
        & Model Parameter Randomization Test (MPRT)& C & \Yes & \No  & \No \\
        \rowcolor{gray}
        & Non-Sensitivity                  & C & \Yes & \No  & \No    \\
        \rowcolor{white}
        & Random Logit Test                & C & \Yes & \No  & \No    \\
        \rowcolor{gray}
        & Smooth MPRT                      & C & \Yes & \No  & \No    \\
        \rowcolor{white}
        & Sparseness                       & C & \Yes & \No  & \No    \\
        \rowcolor{gray}
        \multirow{-10}{*}{\rotatebox[origin=c]{90}{\textbf{Other}}}
        & Sufficiency\tnote{1}~~by Dasgupta et al.~\cite{Dasgupta2022} & C & \Yes & \No  & \No    \\
		\end{xltabular}
\end{ThreePartTable}

\end{document}